\documentclass[runningheads]{llncs}
\usepackage{cite}
\usepackage{amsmath,amssymb,amsfonts}
\usepackage{algorithmic}
\usepackage{graphicx}
\usepackage{textcomp}
\usepackage[table,dvipsnames]{xcolor}
\usepackage[most]{tcolorbox}
\usepackage{booktabs}
\usepackage{tabularx}
\usepackage{xspace}
\usepackage{multirow}
\usepackage{pifont}
\usepackage{enumitem}
\usepackage{graphicx}
\usepackage{caption}
\usepackage{subcaption}
\usepackage[colorlinks,
citecolor=purple,
urlcolor=purple,
linkcolor=purple,
bookmarks=false,
hypertexnames=true]{hyperref}
\usepackage{orcidlink}

\usepackage{multicol}
\def\BibTeX{{\rm B\kern-.05em{\sc i\kern-.025em b}\kern-.08em
    T\kern-.1667em\lower.7ex\hbox{E}\kern-.125emX}}

\newcommand{\mynote}[2]{}

\newif\ifanonymous

\definecolor{cgreen}{rgb}{0.0, 0.42, 0.24}
\definecolor{quitepink}{rgb}{0.858, 0.188, 0.478}
\newcommand{\pname}{\hyphenchar\font=`\-
\texttt{Freddie}\xspace}

\newcommand{\orcidIDL}[1]{\orcidlink{#1}}

\begin{document}
\title{Parameterizing Federated Continual Learning for Reproducible Research}

\author{Bart Cox\orcidIDL{0000-0001-5209-6161} \and
Jeroen Galjaard\orcidIDL{0000-0003-3681-7226} \and
Aditya Shankar\orcidIDL{0009-0009-3046-8724} \and
J\'er\'emie Decouchant\orcidIDL{0000-0001-9143-3984} \and
Lydia Y. Chen\orcidIDL{0000-0002-4228-6735}
}
\authorrunning{B. Cox et al.}
\institute{Delft University of Technology, Delft, The Netherlands\\
\email{\{b.a.cox,j.m.galjaard,a.shankar,j.decouchant\}@tudelft.nl}\\
\email{lydiaychen@ieee.org}}

\maketitle

\begin{abstract}
    Federated Learning (FL) systems evolve in heterogeneous and ever-evolving environments that challenge their performance.
    Under real deployments, the learning tasks of clients can also evolve with time, which calls for the integration of methodologies such as Continual Learning.
    To enable research reproducibility, we propose a set of experimental best practices that precisely capture and emulate complex learning scenarios. Our framework, \pname{}, is the first entirely configurable framework for Federated Continual Learning (FCL), and it can be seamlessly deployed on a large number of machines thanks to the use of Kubernetes and containerization. 
    We demonstrate the effectiveness of \pname{} on two use cases, (i) large-scale FL on CIFAR100 and (ii) heterogeneous task sequence on FCL, which highlight unaddressed performance challenges in FCL scenarios.
\end{abstract}

\section{Introduction}
Federated Learning (FL)~\cite{mcmahan2017communication} performs distributed optimization thanks to a central federator server that maintains a global model using model updates computed by clients. 
It is common for data to be distributed among the clients of an FL system in a non-independent and identically distributed (non-IID) way. 
Moreover, in practice, client learning tasks also evolve over time.
Continual Learning (CL)~\cite{ewc}
is a technique that addresses the scenario where a model is continuously trained on evolving client tasks.

One of the key challenges in CL is catastrophic forgetting: parameters or semantic representations learned for past tasks drift under the influence of new tasks. 
Three categories of techniques address this challenge~\cite{delange}. 
Replay mechanisms, like GEM~\cite{gem} and DGR~\cite{dgr}, retain or generate data from earlier tasks for new task adaptation, which allow the network to revise previously learned tasks.
Regularization techniques, such as EWC~\cite{ewc}, penalize the divergence of model parameters, preventing the adaptation process on new tasks from deviating too far from the model learned on prior tasks. 
Parameter isolation methods use specific weights of the network for the task at hand, i.e., use a mask to freeze the weights of other tasks~\cite{packnet}.

Continual Learning allows a client to learn from its previous tasks if features are repeated over time. 
Federated Continual Learning~\cite{yoon2021federated} (FCL) combines CL and FL, enabling clients to indirectly learn from each other.  
Existing CL frameworks 
do not take this indirect learning into account and therefore 
provide limited support for Federated Continual Learning.

In addition, reproducing FCL results that were obtained in deployment is difficult.
For example, experimental environments are often tightly controlled and steady, while real-world environments are often dynamic and heterogeneous. In addition, clients might be punctually busy processing co-located tasks.
Several FL simulation~\cite{ma2019paddlepaddle,reina2021openfl} and emulation~\cite{beutel2020flower,reina2021openfl} frameworks have been proposed, but they cannot be easily extended to support heterogeneous data, learning tasks and hardware platforms. 
In addition, frameworks that focus on enabling large-scale FL experiments impose a significant overhead to manage the execution or require the use of a strict pipeline.

In this paper, we address the lack of a scalable yet flexible framework for reproducible FCL experiments. 
Overall, we make the following contributions: 
\begin{itemize}[leftmargin=*]
    \item 
    We identify key requirements for FL and FCL emulation: ease of use, reproducibility, support for complex workloads, and resource heterogeneity. 
    \item
    We develop \pname---a framework for \underline{F}ederated and dist\underline{r}ibut\underline{ed} mach\underline{i}ne-l\underline{e}arning---the first open source\footnote{\href{https://gitlab.ewi.tudelft.nl/dmls/publications/freddie}{https://gitlab.ewi.tudelft.nl/dmls/publications/freddie}} 
    framework that addresses these requirements. 
    \pname supports small scale deployments, i.e., single machine simulations, and large scale emulation over self-managed and cloud systems using Kubernetes. \pname enables the emulation of both data and resource heterogeneity. 
    \item
    We provide benchmarking generating methods for FCL that explore both data and task heterogeneity across clients ---realistic workloads tailored for Federated Continual Learning systems.
\end{itemize}

\section{Related Work}

\textbf{Federated Learning (FL)}. Existing FL frameworks support a fixed set of learning tasks across clients.  Flower~\cite{beutel2020flower} provides a client-server framework that needs to be manually started on different devices. 
Differently, Fate~\cite{web:fate} focuses on providing a secure and production-ready federated learning setup.
Fate supports Kubernetes deployments but requires the use of its pipelines to run experiments.
Although this might provide desirable security additions for production, it also tends less to prototyping and active research needs.
Besides research endeavours, popular deep learning platforms such as TorchX 
and Tensorflow Federated
can respectively run distributed and federated experiments at scale, but
they lack the flexibility to use other ML libraries. \\

\textbf{Continual Learning (CL)}. FACIL~\cite{masana2020class}, PyCIL~\cite{zhou2021pycil}, and Pycontinual~\cite{ke2021achieve} provide CL frameworks and CL algorithms such as LwF, iCaRL, EWC, and GEM. 
Continual World~\cite{wolczyk2021continual} adds a simulation world for robotics tasks for Continual Reinforcement Learning.
Avalanche~\cite{lomonaco2021avalanche} is focused on reproducible End-to-End Continual Learning. 
The aforementioned frameworks support CL only on a single machine. 
FedWEIT~\cite{yoon2021federated} combines parameter isolation and regularization and extends CL to a federated setting.
However, it does not consider the impact of task sequences on the global model's quality. 
Last but not least, current FL frameworks cannot be easily extended to support CL scenarios where the output types evolve. 

Continual Learning methods leverage task IDs during training and evaluation, enabling the exploitation of a specific set of model weights or restricting the output classes based on a task's ID. 
This is also known as Task-Interactive Learning (\emph{task-IL}) and Domain-Interactive Learning (\emph{domain-IL})~\cite{threescenarios}, which are both supported by \pname{}.

\section{Specifying Data Heterogeneity and Learning Workloads}

This section briefly surveys system requirements addressed by \pname.

\textbf{Classical FL parameters.} 
The number of clients, the federators' aggregation and client selection strategies must be configurable.  
In addition, FL-related and common hyperparameters, such as the training epochs, learning rates, etc., must be configurable.


\textbf{Statistical heterogeneity.} As in FL, local data distributions remain of high importance, and their non-iidness should be configurable. In the context of FCL, local distributions also limit the tasks that clients might be able to train for. 

\textbf{Resource heterogeneity.} It should be possible to specify the computing power of the clients and of the federator, and the characteristics (latency, throughput) of the network links that interconnect them.

\textbf{Task description.} 
The task sequence of each client can be specified. Non-IID task distribution can be assimilated to the situation where clients learn tasks with high intra-task variance, e.g. due to different domains. In such settings, it is often unclear how the quality of current CL methods is impacted by aggregation.

\label{sec:complex_fcl_workloads}

\begin{figure*}[!tbp]
    \centering
    \includegraphics[width=\linewidth]{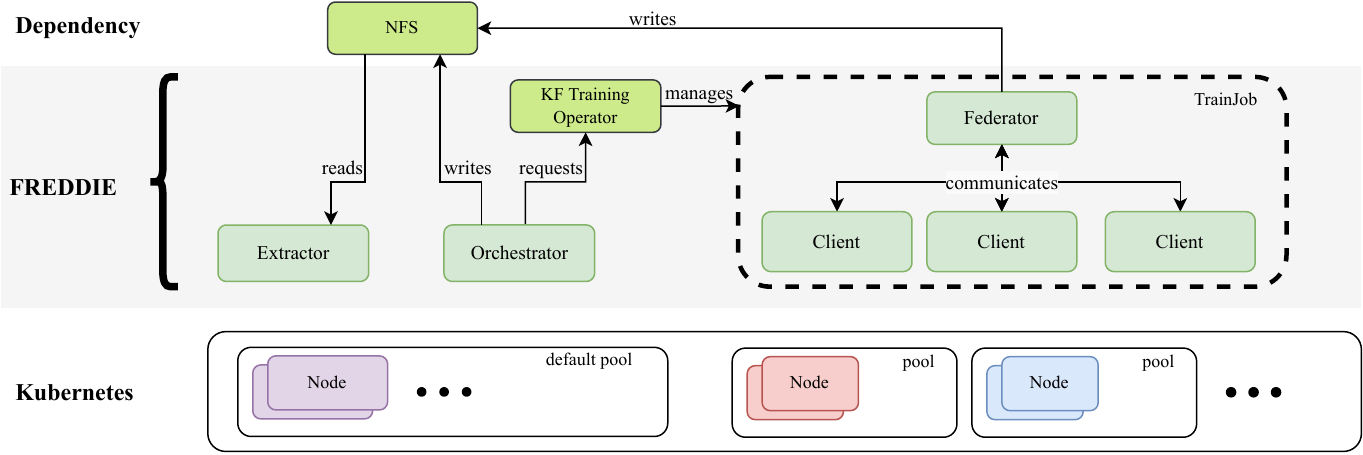}
    \caption{Overview of~\pname{}. An Orchestrator and an Extractor are respectively used for deploying experiments and collecting data.
    Experiments are run as TrainJobs managed by Kubeflow Training Operators.
    Within such a job, the experiment is controlled by the federator and learned by the clients.} 
    \label{fig:kube_stack}
\end{figure*}

\section{\pname: A Framework for Reproducible FCL Research}

In this section, for space reasons, we focus on \pname's implementation based on containerization and orchestration methods, and on its support for FCL. \\

\textbf{Kubernetes and containers.}
Fig.~\ref{fig:kube_stack} presents a high-level overview of \pname. 
The Orchestrator provides functionality to kick-off experiments, in turn, deployed on the cluster by Kubeflow's training operators.
The extractor provides volumes for experiments to write to provided by an NFS provisioner and server.
Experiments themselves are performed by federator and client nodes. The overall flow of the federated learning system with \pname is as follows.
First, the user submits an experiment description of the system and hyper-parameters. 
Following this, the Orchestrator deploys and monitors the experiment.
This design allows the user to scale his experiment up from small-scale prototyping with minimal effort.
The Extractor allows users to store and retrieve experiment statistics and artefacts created by the federator or clients.  

The communication between any two parties in the system is asynchronous, allowing the development of FL systems with non-blocking federator-client interactions.  
With this flexibility, clients can run in Kubernetes clusters and on individual machines. 
To allow users to describe their experiment as distributed code, we rely on Kubeflow's training operators, which provide a means to set up distributed learning on Kubernetes with popular ML libraries. \\

\textbf{Novel support for FCL.}
\pname{} supports the SOTA algorithms for FCL~\cite{yoon2021federated} and common CL methods such as EWC~\cite{ewc} and GEM~\cite{gem}.
For CL, \pname implements Task-IL and Domain-IL~\cite{threescenarios} through the use of sliding, expanding, and full window mechanisms. 
A sliding window restricts the output classes only to those of the task evaluated at a time $t$.
Expanding windows do not utilize the task IDs to make any such restriction, so the output classes include all classes learned until that time. 
A full window does not restrict outputs based on the task and can be used in the standard federated learning scenario. 

\begin{figure}[!tbp]
    \centering
    \begin{subfigure}[T]{0.35\linewidth}
    \centering
        \includegraphics[height=6.2em]{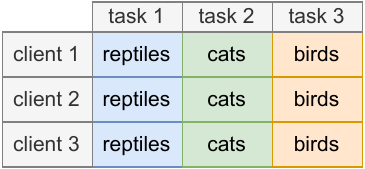}
        \caption{Column-wise.}
        \label{fig:partn_column}
    \end{subfigure}
    \begin{subfigure}[T]{0.3\linewidth}
        \centering\includegraphics[height=6.2em,clip,trim=1.55cm 0cm 0cm 0cm]{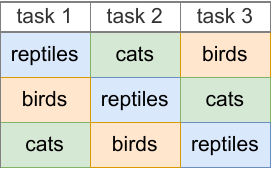}
        \caption{Balanced.}
        \label{fig:partn_bal}
    \end{subfigure}
    \begin{subfigure}[T]{0.3\linewidth}
        \centering\includegraphics[height=6.2em,clip,trim=1.55cm 0cm 0cm 0cm]{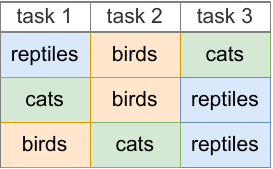}
        \caption{Shuffled.}
        \label{fig:partn_shuff}
    \end{subfigure}
    \caption{Column, balanced and shuffled task partition schemes for CL.}
\end{figure}

The added complexity of FCL allows for workloads over the same set of tasks to produce different results. 
We devise three different schemes that partition tasks differently, and that can be used to evaluate a FCL scheme over a representative set of scenarios.
We discuss these three schemes, which we coin Column, Balanced, and Shuffled, respectively. \textit{Column}, as shown in Fig.~\ref{fig:partn_column}, all clients handle tasks in the same order. 
Thereby naively splitting the CL workload across clients, resulting in a significant expected catastrophic forgetting effect. 
\textit{Balanced}, as depicted in Fig.~\ref{fig:partn_bal}, aims to prevent catastrophic forgetting by organizing the tasks so that different clients train on them across multiple steps. 
We propose a partitioning scheme to lessen the effect of catastrophic forgetting, resulting in a task being trained on by at most one client.
While this scheme addresses short term-forgetting, long-term forgetting may still occur.
\textit{Shuffled}, see Fig.~\ref{fig:partn_shuff}, randomly orders each client's task, thereby relying on pseudo-randomness in conjunction with a pre-specified seed.

\section{Performance evaluation}

We demonstrate some features of \pname{} through experiments. For this purpose, we use the overlapping CIFAR100 dataset. The labels that already exist in CIFAR100 are used to partition the data into different tasks for FCL, following the same steps as in~\cite{yoon2021federated}. 
FL can be viewed as a corner case of FCL where there is a single task consisting of the whole dataset. 
We first consider a FL scenario with the default version of CIFAR100, and then consider the overlapping CIFAR100 split into 10 separate tasks in a FCL scenario. We use the average accuracy metric following the CL literature~\cite{chaudhry2018efficient, mirzadeh2020understanding}.\\

\textbf{Scalability.} To investigate \pname's emulation capability, we perform a \emph{small} and \emph{large} scale experiment on a Google Kubernetes Engine (GKE) cluster to cover possible use cases.
During deployment, the pods of the federation and clients were run on a separate node pool scaled to meet each experiment's requirements.
We study the performance of an FL experiment emulated on a CPU-enabled Kubernetes cluster, where multiple clients may run on a single Kubernetes node.
Parameters of the experiments are provided in Tab.~\ref{tab:config_freddie_kube}.

\begin{figure}[tbp]
    \centering
    \begin{subfigure}[T]{0.48\linewidth}
        \centering
        \includegraphics[width=\linewidth]{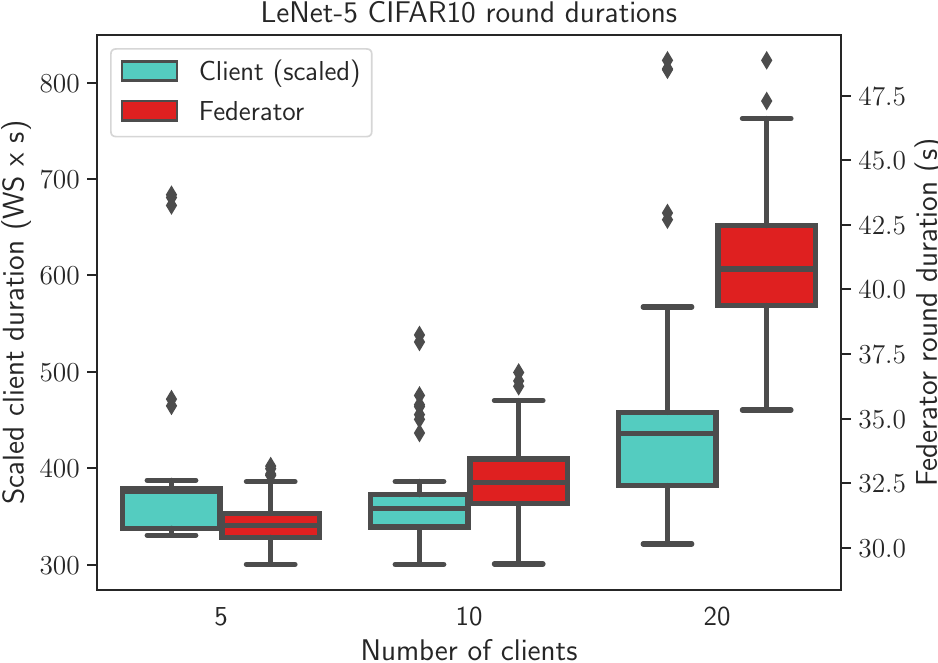}
        \caption{Experiments with 5 participating clients per round (n=1).}
        \label{fig:freddie_kube_small}
    \end{subfigure}
    ~
    \begin{subfigure}[T]{0.48\linewidth}
        \centering
        \includegraphics[width=\linewidth]{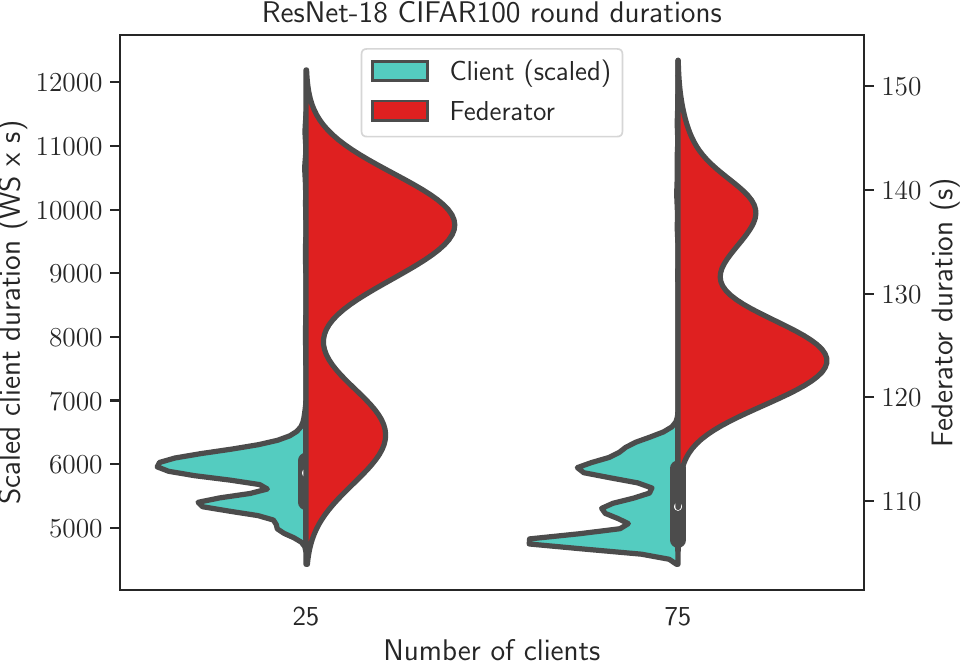}
        \caption{Scaled experiments where all clients participate in each round (n=3).}
        \label{fig:freddie_kube_scale}
    \end{subfigure}
    \caption{Client and federator round durations with \pname for small (5-20 clients, LeNet5 \& CIFAR10) and large scale experiment (ResNet-18 \& CIFAR100). Client durations are scaled by the total number of clients (WS).
    }
    \label{fig:freddie_kube}
\end{figure}

\begin{table}[!t]
\caption{System and hyperparameters used in `small' and `scale' experiments. All experiments were run on `e2-standard-8' nodes.}
    \label{tab:config_freddie_kube}
\begin{tabular}{@{}lcccccccccc@{}}
\toprule
      & \multicolumn{4}{c}{System}                                                                                                      & \multicolumn{3}{c}{Federator (F)}                                                        & \multicolumn{3}{c}{Clients (C)}          \\ \cmidrule(l){2-11} 
      & Nodes & \#C     & \begin{tabular}[c]{@{}c@{}}CPU\\  (F/C)\end{tabular} & \begin{tabular}[c]{@{}c@{}}Memory\\ (F/C)\end{tabular} & Strategy                & \begin{tabular}[c]{@{}c@{}}\#Rounds\\ (R)\end{tabular} & \#C/R & Model  & Data     & BS                  \\ \midrule
Small & 2,2,3 & 5,10,20 & \multirow{2}{*}{2/1}                                 & 2/2G                                                   & \multirow{2}{*}{FedAvg} & 100                                                    & 5     & LeNet  & CIFAR10  & \multirow{2}{*}{64} \\
Scale & 4,12  & 25,75   &                                                      & 2/2,6G                                                 &                         & 85                                                     & all   & ResNet & CIFAR100 &                     \\ \bottomrule
\end{tabular}
                        
\end{table}
The \emph{small} experiment in Fig.~\ref{fig:freddie_kube_small} depicts the spread round times of clients (scaled) and the federator, with 5 selected clients per round.
The client round duration is scaled by the number of clients (World Size WS) ($|\mathcal{D}_{Cifar}|/\text{WS}$) to account for differences in clients' datasets as the WS increases.
The outliers in the plots originate from the first epoch run on clients, which are inherently slower due to loading data into memory.
Nevertheless, it is expected that the scaled client duration stays relatively constant, while the result shows an increase as the number of clients increases (from 115 to {123\,s}, and from 136 to {138\,s}).
Similarly, the federator sees a positive correlation between round duration and WS.
The number of co-scheduled clients on the same node can explain this trend, as the networking overhead stays the same.  

For the \textit{scale} experiments, we provide the round time density estimate in Fig.~\ref{fig:freddie_kube_scale}. 
The client round times exhibit the same range of processing times that were observed in the `small' setting.
In both settings, participating clients in each round may run on the same node, varying from 3 to 7 clients per node.
We use similar settings in the `small' configuration that involves 20 clients, where 4 nodes are used. 
As such, confirming that resource contingencies due to co-scheduling will likely cause the increased client round time with 20 clients. 
The different modes within the client's round duration can be explained by imperfect data splits and the imbalanced assignment of the number of clients to be co-scheduled with the federator. 
The federator's density estimate shows a similar pattern with two distinct modes.
With the cluster configurations employed, i.e., 4 and 12 nodes, it is possible for the federator to be co-scheduled on a machine with different numbers of clients.
As a result, the federator experiences a variable level of resource contingency.
However, an increase in the two modes is visible as the number of clients increases, which is expected due to the increased communication volumes. \\

\begin{figure*}[!t]
\begin{subfigure}[T]{0.55\linewidth}
        \centering
        \includegraphics[height=10em]{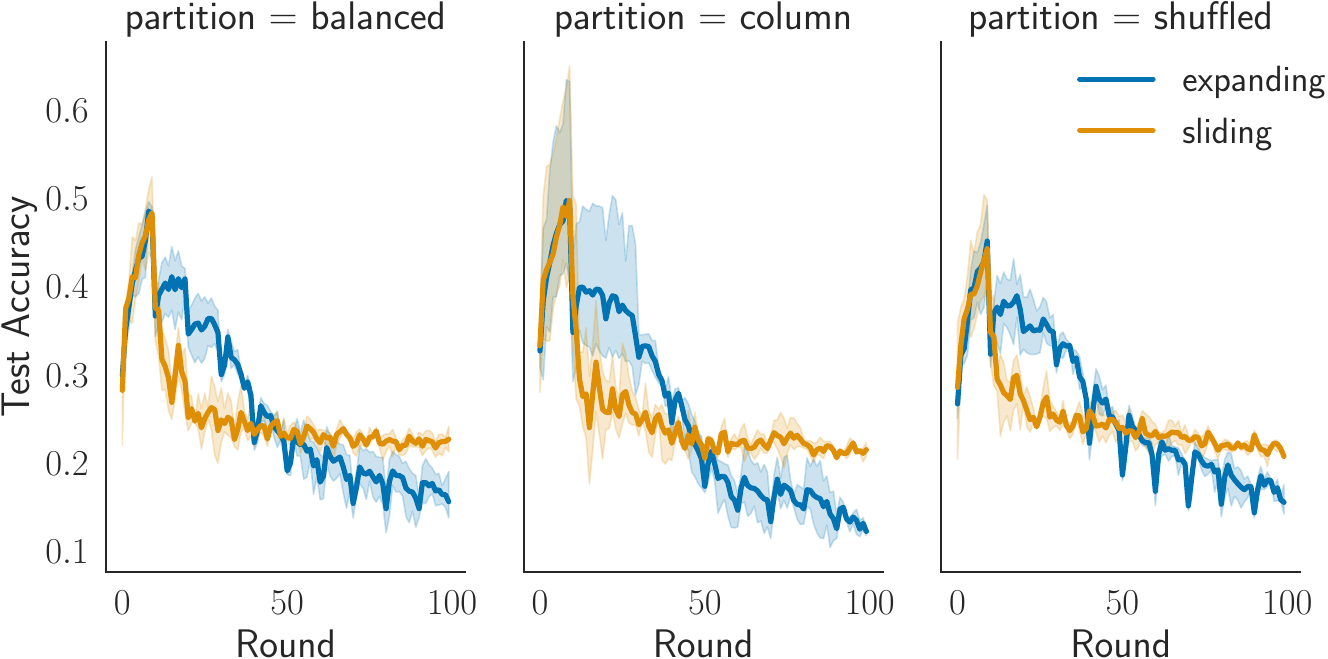}
        \caption{Reduced catastrophic forgetting effect window (Task-IL) compared to expanding window (Domain-IL). 
        }
        \label{fig:use_case_aditya_2}
\end{subfigure}
~
\begin{subfigure}[T]{0.4\linewidth}
    \centering
        \centering
        \includegraphics[height=10em]{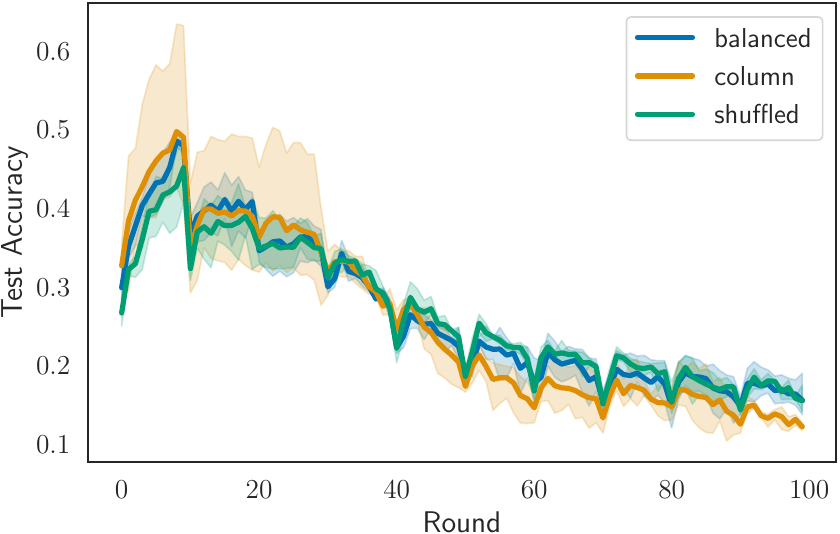}
        \caption{Impact of partition scheme on evaluation accuracy (10 tasks of overlapping CIFAR100, LeNet-5, 10 rounds of 2 epochs per task).}
        \label{fig:use_case_aditya_1}
    \end{subfigure}
        \caption{Impact of task heterogeneity on FCL.}
\end{figure*}

\textbf{Task-IL vs Domain-IL.}
Assuming that the model knows the ID of the task it is currently training on, or evaluating, increases its accuracy. 
For FCL, Task-Interactive Learning and Domain-Interactive Learning are implemented using the sliding and expanding-window, respectively. 
Let us recall that sliding-windows use task IDs, contrary to expanding-windows.
For the overlapping CIFAR100 dataset, if one assumes that the task ID is known, then the number of output classes is restricted to only the five sub-classes within that task.
Thus leading to higher average task probabilities for Task-IL scenarios.
This difference is prevalent in Fig.~\ref{fig:use_case_aditya_2}.
Under the expanding window scheme, classification outputs one of $5T$ classes, where $T$ is the number of tasks learned until evaluation time. 
Therefore, the probability of classifying correctly is even lower than in the sliding window scenario. 
Fig.~\ref{fig:use_case_aditya_2} shows the positive impact of using a task ID on accuracy.
Using sliding-window results in higher accuracy than expanding-window, which sometimes has to be used because of the application use case. 
Because of this difference, \pname supports both Task-IL and Domain-IL.

\noindent

\textbf{FCL Task Heterogeneity.}
As discussed in Section~\ref{sec:complex_fcl_workloads}, tasks can be processed in different orders at each client. 
To demonstrate the different effects that different sequences of tasks produce, we implement the Overlapped-CIFAR100 dataset with 20 tasks that can be used for FCL~\cite{yoon2021federated}.
The accuracy in Fig.~\ref{fig:use_case_aditya_1} is calculated as the average accuracy of all tasks seen until that point, resulting in expected `drops' in accuracy as new tasks are introduced. 
Indeed, the learning curves in Fig.~\ref{fig:use_case_aditya_1} show noticeable drops over time. 
However, different trends are visible between workloads. 
The column scheme suffers more from more pronounced \emph{catastrophic forgetting} than the shuffled and balanced scheme, resulting in lower accuracy.
We observe that the column scheme, on average, results in a 4\% test accuracy drop compared to the column and shuffled schemes. 

\section{Conclusion}
We presented \pname, the first framework for reproducible Federated Continual Learning research, which is motivated by the increasing importance of Federated and Continual Learning. 
\pname's deployment abilities on different platforms, scalability with the number of clients, and support for data and task heterogeneity provide FL practitioners with a powerful tool.
Our experimental results showcase previously unaddressed performance issues that Federated Continual Learning systems might face: severe catastrophic forgetting in different task heterogeneity settings. 
\pname is open-source, and will soon be extended to support new CL datasets, algorithms, and generative models.

\bibliographystyle{splncs04}
\bibliography{refs-short}
\end{document}